\documentclass{article}
\usepackage{ijcai18}

\usepackage{times}
\usepackage{xcolor}
\usepackage{soul}
\usepackage[utf8]{inputenc}
\usepackage[small]{caption}

\usepackage{graphicx}
\usepackage{algorithm}
\usepackage{amssymb}
\usepackage{amsmath}
\usepackage{mathrsfs}
\usepackage{algpseudocode}
\usepackage{tabularx}
\usepackage{multicol}
\usepackage{booktabs}
\usepackage{subfigure}
\usepackage{verbatim}
\usepackage{xcolor}
\usepackage{verbatim}

\newtheorem{theorem}{Theorem}
\newtheorem{corollary}{Corollary}[theorem]

\title{Request-and-Reverify: Hierarchical Hypothesis Testing for \\ Concept Drift Detection with Expensive Labels}

\author{
Shujian Yu\thanks{This work was done when Shujian Yu was a research intern at Nokia Bell Labs, Murray Hill, NJ, USA.}$^{1,2}$,
Xiaoyang Wang$^1$,
Jos\'{e} C. Pr\'{i}ncipe$^2$
\\
$^1$ Nokia Bell Labs, Murray Hill, NJ, USA\\
$^2$ Dept. of Electrical and Computer Engineering, University of Florida, Gainesville, FL, USA\\
yusjlcy9011@ufl.edu,
xiaoyang.wang@nokia-bell-labs.com,
principe@cnel.ufl.edu
}

\providecommand{\ve}[1]{\boldsymbol{\mathrm{#1}}}

\algdef{SE}[VARIABLES]{Variables}{EndVariables}
   {\algorithmicvariables}
   {\algorithmicend\ \algorithmicvariables}
\algnewcommand{\algorithmicvariables}{\textbf{variables declaration}}

\begin{document}

\maketitle

\begin{abstract}
One important assumption underlying common classification models is the stationarity of the data. However, in real-world streaming applications, the data concept indicated by the joint distribution of feature and label is not stationary but drifting over time. Concept drift detection aims to detect such drifts and adapt the model so as to mitigate any deterioration in the model's predictive performance. Unfortunately, most existing concept drift detection methods rely on a strong and over-optimistic condition that the true labels are available immediately for all already classified instances. In this paper, a novel Hierarchical Hypothesis Testing framework with \textbf{Request-and-Reverify} strategy is developed to detect concept drifts by requesting labels only when necessary. Two methods, namely Hierarchical Hypothesis Testing with Classification Uncertainty (HHT-CU) and Hierarchical Hypothesis Testing with Attribute-wise ``Goodness-of-fit'' (HHT-AG), are proposed respectively under the novel framework. In experiments with benchmark datasets, our methods demonstrate overwhelming advantages over state-of-the-art unsupervised drift detectors. More importantly, our methods even outperform DDM (the widely used supervised drift detector) when we use significantly fewer labels.
\end{abstract}

\section{Introduction}
\noindent In the last decades, numerous efforts have been made in algorithms that can learn from data streams. Most traditional methods for this purpose assume the stationarity of the data. However, when the underlying source generating the data stream, i.e., the joint distribution $\mathbb{P}_t(\ve{X},y)$, is not stationary, the optimal decision rule should change over time. This is a phenomena known as concept drift~\cite{ditzler2015learning,krawczyk2017ensemble}. Detecting such concept drifts is essential for the algorithm to adapt itself to the evolving data.

Concept drift can manifest two fundamental forms of changes from the Bayesian perspective \cite{kelly1999impact}: 1) a change in the marginal probability $\mathbb{P}_t(\ve{X})$; 2) a change in the posterior probability $\mathbb{P}_t(y|\ve{X})$. Existing studies in this field primarily concentrate on detecting posterior distribution change $\mathbb{P}_t(y|\ve{X})$, also known as the real drift~\cite{widmer1993effective}, as it clearly indicates the optimal decision rule. On the other hand, only a little work aims at detecting the virtual drift \cite{hoens2012learning}, which only affects $\mathbb{P}_t(\ve{X})$. In practice, one type of concept drift typically appears in combination with the other~\cite{tsymbal2004problem}. Most methods for real drift detection assume that the true labels are available immediately after the classifier makes a prediction. However, this assumption is over-optimistic, since it could involve the annotation of data by expensive means in terms of cost and labor time. The virtual drift detection, though making no use of true label $y_t$, has the issue of wrong interpretation (i.e., interpreting a virtual drift as the real drift). Such wrong interpretation could provide wrong decision about classifier update which still require labeled data~\cite{krawczyk2017ensemble}.

To address these issues simultaneously, we propose a novel Hierarchical Hypothesis Testing (HHT) framework with a \textbf{Request-and-Reverify} strategy for concept drift detection. HHT incorporates two layers of hypothesis tests. Different from the existing HHT methods~\cite{alippi2017hierarchical,yu2017concept}, our HHT framework is the first attempt to use labels for concept drift detection \textbf{only when necessary}. It ensures that the test statistic (derived in a fully unsupervised manner) in Layer-I captures the most important properties of the underlying distributions, and adjusts itself well in a more powerful yet conservative manner that only requires labeled data when necessary in Layer-II. Two methods, namely Hierarchical Hypothesis Testing with Classification Uncertainty (HHT-CU) and Hierarchical Hypothesis Testing with Attribute-wise ``Goodness-of-fit'' (HHT-AG), are proposed under this framework in this paper. The first method incrementally tracks the distribution change with the defined $\emph{classification uncertainty}$ measurement in Layer-I, and uses permutation test in Layer-II, whereas the second method uses the standard Kolmogorov-Smirnov (KS) test in Layer-I and two-dimensional ($2$D) KS test \cite{peacock1983two} in Layer-II. We test both proposed methods in benchmark datasets. Our methods demonstrate overwhelming advantages over state-of-the-art unsupervised methods. Moreover, though using significantly fewer labels, our methods outperform supervised methods like DDM~\cite{gama2004learning}.

\section{Background Knowledge} \label{section2}
\subsection{Problem Formulation}
Given a continuous stream of labeled samples $\{\ve{X}_t,y_t\}$, $t=1,2,...,T$, a classification model $\hat{f}$ can be learned so that $\hat{f}(\ve{X}_t) \mapsto y_t$. Here, $\ve{X}_t \in \mathbb{R}^d$ represents a $d$-dimensional feature vector, and $y_t$ is a discrete class label. Let $(\ve{X}_{T+1},\ve{X}_{T+2},...,\ve{X}_{T+N})$ be a sequence of new samples that comes chronologically with unknown labels. At time $T+N$, we split the samples in a set $S_A=(\ve{X}_{T+N-n_A+1},\ve{X}_{T+N-n_A+2},...,\ve{X}_{T+N})$ of $n_A$ recent ones and a set $S_B=(\ve{X}_{T+1},\ve{X}_{T+2},...,\ve{X}_{T+N-n_A})$ containing the $(N-n_A)$ samples that appear prior to those in $S_A$. The problem of \emph{concept drift detection} is identifying whether or not the source $\mathscr{P}$ (i.e., the joint distribution $\mathbb{P}_t(\ve{X},y)$\footnote{The distributions are deliberated subscripted with time index $t$ to explicitly emphasize their time-varying characteristics.}) that generates samples in $S_A$ is the same as that in $S_B$ (even without access to the true labels $y_t$)~\cite{ditzler2015learning,krawczyk2017ensemble}. Once such a drift is found, the machine can request a window of labeled data to update $\hat{f}$ and employ the new classifier to predict labels of incoming data.

\subsection{Related Work}
The techniques for concept drift detection can be divided into two categories depending on reliance of labels~\cite{sethi2017reliable}: supervised (or explicit) drift detectors and unsupervised (or implicit) drift detectors. \textbf{Supervised Drift Detectors} rely heavily on true labels, as they typically monitor one error metrics associated with classification loss. Although much progress has been made on concept drift detection in the supervised manner, its assumption that the ground truth labels are available immediately for all already classified instances is typically over-optimistic. \textbf{Unsupervised Drift Detectors}, on the other hand, explore to detect concept drifts without using true labels. Most unsupervised concept drift detection methods concentrate on performing multivariate statistical tests to detect the changes of feature values $\mathbb{P}_t(\ve{X})$, such as the Conjunctive Normal Form (CNF) density estimation test~\cite{dries2009adaptive} and the Hellinger distance based density estimation test~\cite{ditzler2011hellinger}. Considering their high computational complexity, an alternative approach is to conduct univariate test on each attribute of features independently. For example, \cite{reis2016fast} develops an incremental (sequential) KS test which can achieve exactly the same performance as the conventional batch-based KS test.

Besides modeling virtual drifts of $\mathbb{P}_t(\ve{X})$, recent research in unsupervised drift detection attempts to model the real drifts by monitoring the classifier output $\hat{y}_t$ or posterior probability as an alternative to $y_t$. The Confidence Distribution Batch Detection (CDBD) approach~\cite{lindstrom2011drift} uses Kullback-Leibler (KL) divergence to compare the classifier output values from two batches. A drift is signaled if the divergence exceeds a threshold. This work is extended in \cite{kim2017efficient} by substituting the classifier output value with the classifier confidence measurement. Another representative method is the Margin Density Drift Detection (MD3) algorithm~\cite{sethi2017reliable}, which tracks the proportion of samples that are within a classifier (i.e., SVM) margin and uses an active learning strategy in \cite{zliobaite2014active} to interactively query the information source to obtain true labels. Though not requiring true labels for concept drift detection, the major drawback of these unsupervised drift detectors is that they are prone to false positives as it is difficult to distinguish noise from distribution changes. Moreover, the wrong interpretation of virtual drifts could cause wrong decision for classifier update which require not only more labeled data but also unnecessary classifier re-training~\cite{krawczyk2017ensemble}.

\section{Request-and-Reverify HHT Approach}

\begin{figure}
\centering
\includegraphics[height=3.5cm,width=8.5cm]{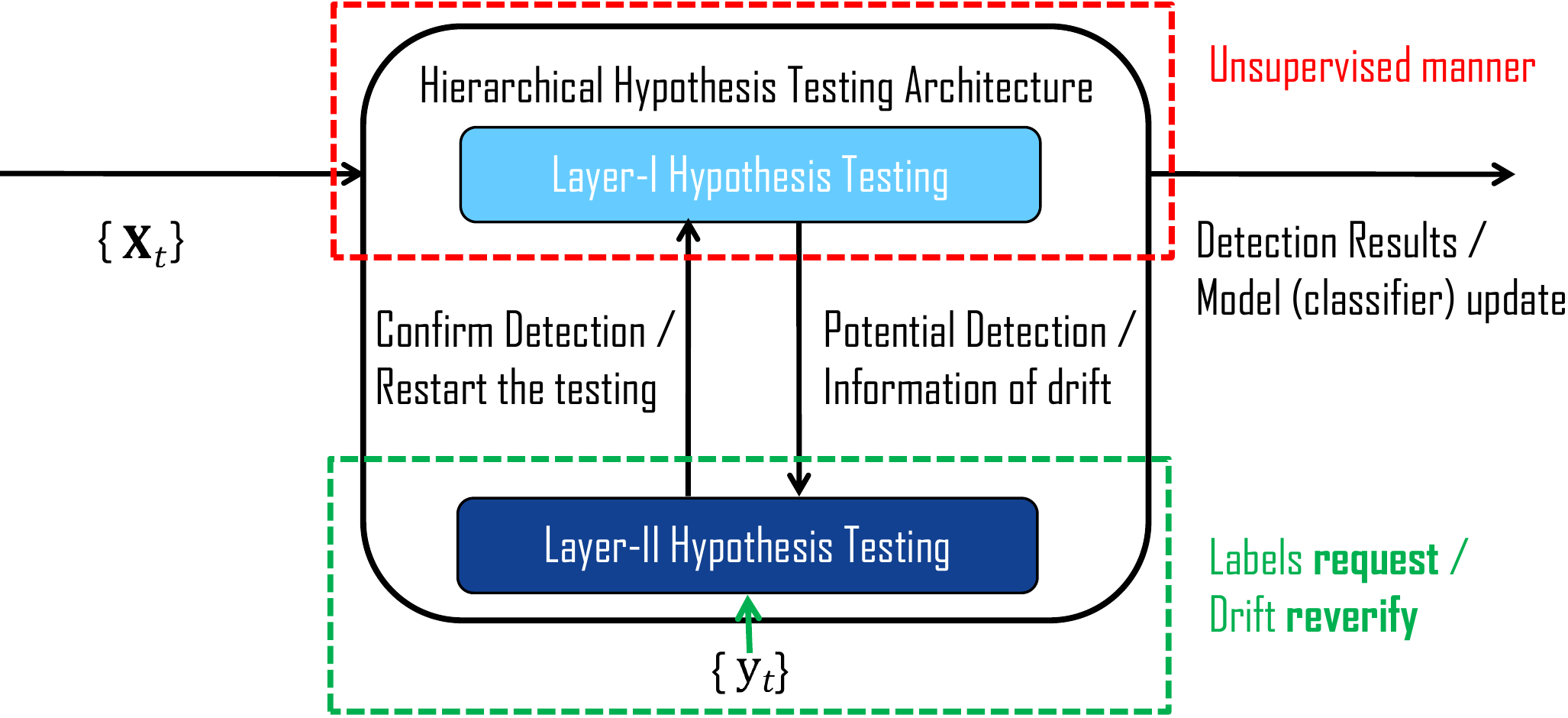}
\caption{The Request-and-Reverify Hierarchical Hypothesis Testing framework for concept drift detection with expensive labels.}
\label{fig:hierarchical architecture}
\end{figure}

The observations on the existing supervised and unsupervised concept drift detection methods motivate us to propose the Request-and-Reverify Hierarchial Hypothesis Testing framework (see Fig.~\ref{fig:hierarchical architecture}). Specifically, our layer-I test is operated in a fully unsupervised manner that does not require any labels. Once a potential drift is signaled by Layer-I, the Layer-II test is activated to confirm (or deny) the validity of the suspected drift. The result of the Layer-II is fed back to the Layer-I to reconfigure or restart Layer-I once needed.%

In this way, the upper bound of HHT's Type-I error is determined by the significance level of its Layer-I test, whereas the lower bound of HHT's Type-II error is determined by the power of its Layer-I test. Our Layer-I test (and most existing single layer concept drift detectors) has low Type-II error (i.e., is able to accurately detect concept drifts), but has relatively higher Type-I error (i.e., is prone to generate false alarms). The incorporation of the Layer-II test is supposed to reduce false alarms, thus decreasing the Type-I error. The cost is that the Type-II error could be increased at the same time. In our work, we request true labels to conduct a more precise Layer-II test, so that we can significantly decrease the Type-I error with minimum increase in the Type-II error.

\subsection{HHT with Classification Uncertainty (\small{HHT-CU})}
Our first method, HHT-CU, detects concept drift by tracking the $\emph{classification uncertainty}$ measurement $u_t=\|\hat{y}_t-\hat{\mathbb{P}}(y_t|\ve{X}_t)\|_2$, where $\|\cdot\|_2$ denotes the $\ell_2$ distance, $\hat{\mathbb{P}}(y_t|\ve{X}_t)$ is the posterior probability estimated by the classifier at time index $t$, and $\hat{y}_t$ is the target label encoded from $\hat{\mathbb{P}}(y_t|\ve{X}_t)$ using the $1$-of-$K$ coding scheme~\cite{bishop2006pattern}. Intuitively, the distance between $\hat{y}_t$ and $\hat{\mathbb{P}}(y_t|\ve{X}_t)$ measures the $\emph{classification uncertainty}$ for the current classifier, and the statistic derived from this measurement should be stationary (i.e., no ``significant'' distribution change) in a stable concept. Therefore, the dramatic change of the uncertainty mean value may suggest a potential concept drift.

Different from the existing work that typically monitors the derived statistic with the three-sigma rule in statistical process control~\cite{montgomery2009introduction}, we use the Hoeffding's inequality \cite{hoeffding1963probability} to monitor the moving average of $u_t$ in our Layer-I test.

\begin{theorem}[Hoeffding's inequality]
Let $X_1$, $X_2$,..., $X_n$ be independent random variables such that $X_i\in[a_i,b_i]$, and let $\bar{X}=\frac{1}{n}\sum^{n}_{i=1}X_i$, then for $\varepsilon\geq0$:
\begin{equation}
\mathbb{P}\{\bar{X}-\mathbb{E}(\bar{X})\geq\varepsilon\}\leq e^{\frac{-2n^2\varepsilon^2}{\sum^{n}_{i=1}(b_i-a_i)^2}}.
\end{equation}
\end{theorem}
where $\mathbb{E}$ denotes the expectation. Using this theorem, given a specific significance level $\alpha$, the error $\varepsilon_{\alpha}$ can be computed as:
\begin{equation}\label{error_X}
\varepsilon_{\alpha}=\sqrt{\frac{1}{2n}\ln{\frac{1}{\alpha}}}.
\end{equation}

The Hoeffding's inequality does not require an assumption on the probabilistic distribution of $u_t$. This makes it well suited in learning from real data streams \cite{frias2015online}. Moreover, the Corollary~\ref{theorem:corrol} proposed by Hoeffding \cite{hoeffding1963probability} can be directly applied to detect significant changes in the moving average of streaming values.

\begin{corollary}[Layer-I test of HHT-CU]\label{theorem:corrol}
If $X_1$, $X_2$, ..., $X_n$, $X_{n+1}$, ..., $X_{n+m}$ be independent random variables with values in the interval $[a,b]$, and if $\bar{X}=\frac{1}{n}\sum^{n}_{i=1}X_i$ and $\bar{Z}=\frac{1}{n+m}\sum^{n+m}_{i=1}X_i$, then for $\varepsilon\geq0$:
\begin{equation}
\mathbb{P}\{\bar{X}-\bar{Z}-(\mathbb{E}(\bar{X})-\mathbb{E}(\bar{Z}))\geq\varepsilon\}\leq e^{\frac{-2n(n+m)\varepsilon^2}{m(b-a)^2}}.
\end{equation}
\end{corollary}

By definition, $u_t\in[0,\sqrt{\frac{K-1}{K}}]$, where $K$ is the number of classes. $\bar{X}$ denotes the \emph{classification uncertainty} moving average before a cutoff point, and $\bar{Z}$ denotes the moving average over the whole sequence. The rule to reject the null hypothesis $H_0: \mathbb{E}(\bar{X})> \mathbb{E}(\bar{Z})$ against the alternative one $H_1: \mathbb{E}(\bar{X})\leq\mathbb{E}(\bar{Z})$ at the significance level $\alpha$ will be $\bar{Z}-\bar{X}\geq\varepsilon_{\alpha}$, where
\begin{equation}\label{error_Z}
\varepsilon_{\alpha}=\sqrt{\frac{K-1}{K}}\times\sqrt{\frac{m}{2n(n+m)}\ln{\frac{1}{\alpha}}}.
\end{equation}

Regarding the cutoff point, a reliable location can be estimated from the minimum value of $\bar{X}_i+\varepsilon_{\bar{X}_i}$ ($1\leq i\leq n+m$)~\cite{gama2004learning,frias2015online}. This is because $\bar{X}_i$ keeps approximately constant in a stable concept, thus $\bar{X}_i+\varepsilon_{\bar{X}_i}$ must reduce its value correspondingly.

The Layer-II test aims to reduce false positives signaled by Layer-I. Here, we use the permutation test which is described in \cite{yu2017concept}. Different from \cite{yu2017concept}, which trains only one classifier $f_{ord}$ using $S_{ord}$ and evaluates it on $S'_{ord}$ to get a zero-one loss $\hat{E}_{ord}$, we train another classifier $f'_{ord}$ using $S'_{ord}$ and evaluate it on $S_{ord}$ to get another zero-one loss $\hat{E}'_{ord}$. We reject the null hypothesis if either $\hat{E}_{ord}$ or $\hat{E}'_{ord}$ deviates too much from the prediction loss of the shuffled splits. The proposed HHT-CU is summarized in Algorithm \ref{HHT-CU}, where the window size $N$ is set as the number of labeled samples to train the initial classifier $\hat{f}$.

\begin{algorithm}[htb]
\small
\caption{\small{HHT with Classification Uncertainty (HHT-CU)}}
\label{HHT-CU}
\begin{algorithmic}[1]
\Require
Unlabeled stream $\{\mathbf{X}_t\}_{t=0}^\infty$ where $\mathbf{X}_t\in \mathbb{R}^d$; Initially trained classifier $\hat{f}$; Layer-I significance level $\Theta_1$; Layer-II significance level $\Theta_2$; Window size $N$.
\Ensure
Detected drift time index $\{T_{cd}\}$; Potential drift time index $\{T_{pot}\}$.
\Variables
 \State $\bar{X}_{cut}$: moving average of $u_1$, $u_2$, ..., $u_{cut}$;
 \State $\bar{Z}_n$: moving average of $u_1$, $u_2$, ..., $u_n$;
 \State $\varepsilon_{\bar{X}_{cut}}$ and $\varepsilon_{\bar{Z}_n}$: error bounds computed using Eqs. (\ref{error_X}) and (\ref{error_Z}) respectively;
\EndVariables

\State $\{T_{cd}\}=\phi$; $\{T_{pot}\}=\phi$;
\For {$t = 1$ to $\infty$}
\State Compute $u_t$ using $\hat{f}$;
\State Update $\bar{Z}_t$ and $\varepsilon_{\bar{Z}_t}$ by adding $u_t$;

\If {$\bar{Z}_t+\varepsilon_{\bar{Z}_t}\leq \bar{X}_{cut}+\varepsilon_{\bar{X}_{cut}}$}
\State $\bar{X}_{cut}=\bar{Z}_t$; $\varepsilon_{\bar{X}_{cut}}=\varepsilon_{\bar{Z}_t}$;
\EndIf

\If {$H_0: \mathbb{E}(\bar{X}_{cut})\geq\mathbb{E}(\bar{Z}_t)$ is rejected at $\Theta_1$}
\State $\{T_{pot}\}\leftarrow t$;
\State Request $2N$ labeled samples $\{\mathbf{X}_i,y_i\}_{i=t-N}^{t+N-1}$;
\State Perform Layer-II test using $\{\mathbf{X}_i,y_i\}_{i=t-N}^{t+N-1}$ at $\Theta_2$;
\If {(Layer-II confirms the potentiality of $t$)}
\State $\{T_{cd}\}\leftarrow t$;
\State Update $\hat{f}$ using $\{\mathbf{X}_i,y_i\}_{i=t}^{t+N-1}$;
\State Initialize declared variables;
\Else
\State Discard $t$;
\State Restart Layer-I test;
\EndIf
\EndIf
\EndFor
\end{algorithmic}
\normalsize
\end{algorithm}

\subsection{HHT with Attribute-wise ``Goodness of fit'' (HHT-AG))}
The general idea behind HHT-AG is to explicitly model $\mathbb{P}_t(\ve{X},y)$ with limited access to $y$. To this end, a feasible solution is to detect potential drift points in Layer-I by just modeling $\mathbb{P}_t(\ve{X})$, and then require limited labeled data to confirm (or deny) the suspected time index in Layer-II. 

The Layer-I test of HHT-AG conducts ``Goodness-of-fit'' test on each attribute $x^k|_{k=1}^d$ individually to determine whether $\ve{X}$ from two windows differ: a baseline (or reference) window $W_1$ containing the first $N$ items of the stream that occur after the last detected change; and a sliding window $W_2$ containing $N$ items that follow $W_1$. We slide the $W_2$ one step forward whenever a new item appears on the stream. A potential concept drift is signaled if at least for one attribute there is a distribution change. Factoring $\mathbb{P}_t(\ve{X})$ into $\prod_{k=1}^d\mathbb{P}_t(x^k)$ for multivariate change detection is initially proposed in \cite{kifer2004detecting}. Since then, this factorization strategy becomes widely used~\cite{vzliobaite2010change,reis2016fast}. Sadly, no existing work provides a theoretical foundation of this factorization strategy. In our perspective, one possible explanation is the Sklar's Theorem~\cite{Sklar1959Fonctions}, which states that if $\mathbb{H}$ is a $d$-dimensional joint distribution function and if $\mathbb{F}_1$, $\mathbb{F}_2$, ..., $\mathbb{F}_d$ are its corresponding marginal distribution functions, then there exists a $d$-copula $C$: $[0,1]^d\rightarrow[0,1]$ such that:
\begin{equation}
\mathbb{H}(\mathbf{X})=C(\mathbb{F}_1(x^1),\mathbb{F}_2(x^2),...,\mathbb{F}_d(x^d)).
\end{equation}
The density function (if exists) can thus be represented as:
\small
\begin{equation}
\mathbb{P}(\mathbf{X})=c(\mathbb{F}_1(x^1),\mathbb{F}_2(x^2),...,\mathbb{F}_d(x^d))\prod_{k=1}^d\mathbb{P}(x^k)\varpropto\prod_{k=1}^d\mathbb{P}(x^k),\nonumber
\end{equation}
\normalsize
where $c$ is the density of the copula $C$.

Though Sklar does not show practical ways on how to calculate $C$, this Theorem demonstrates that if $\mathbb{P}(\ve{X})$ changes, we can infer that one of $\mathbb{P}(x_i)$ should also changes; otherwise, if none of the $\mathbb{P}(x_i)$ changes, the $\mathbb{P}(\ve{X})$ would not be likely to change.

This paper selects Kolmogorov-Smirnov (KS) test to measure the discrepancy of $\mathbb{P}_t(x^k)|_{k=1}^d$ in two windows. Specifically, the KS test rejects the null hypothesis, i.e., the observations in sets $\mathbf{A}$ and $\mathbf{B}$ originate from the same distribution, at significance level $\alpha$ if the following inequality holds:
\begin{equation}
\sup\limits_{x} |\mathbb{F}_\mathbf{A}(x)-\mathbb{F}_\mathbf{B}(x)|>s(\alpha)\sqrt{\frac{m+n}{mn}},
\end{equation}
where $\mathbb{F}_\mathbf{C}(x)=\frac{1}{|\mathbf{C}|}\sum\mathbf{1}_{\{c\in\mathbf{C},c\leq x\}}$ denotes the empirical distribution function (an estimation to the cumulative distribution function $\mathbb{P}(X<x)$), $s(\alpha)$ is a $\alpha$-specific value that can be retrieved from a known table, $m$ and $n$ are the cardinality of set $\mathbf{A}$ and set $\mathbf{B}$ respectively.

We then validate the potential drift points by requiring true labels of data that come from $W_1$ and $W_2$ in Layer-II. The Layer-II test of HHT-AG makes the conditionally independent factor assumption \cite{bishop2006pattern} (a.k.a. the ``naive Bayes'' assumption), i.e., $\mathbb{P}(x^i|x^j,y)=\mathbb{P}(x^i|y)$ ($1\leq i\neq j\leq d$). Thus, the joint distribution $\mathbb{P}_t(\ve{X},y)$ can be represented as:
\small
\begin{align}
\mathbb{P}_t(\ve{X},y)&=\mathbb{P}_t(y)\mathbb{P}_t(x^d|y)\mathbb{P}_t(x^{d-1}|x^d,y)...\mathbb{P}_t(x^1|x^2,...,x^d,y) \nonumber\\
&\varpropto \mathbb{P}_t(y)\prod_{k=1}^d\mathbb{P}_t(x^k|y)\varpropto\prod_{k=1}^d\mathbb{P}_t(x^k,y).
\label{Eq_nB}
\end{align}
\normalsize

According to Eq.~(\ref{Eq_nB}), we perform $d$ independent two-dimensional (2D) KS tests~\cite{peacock1983two} on each bivariate distribution $\mathbb{P}_t(x^k,y)|_{k=1}^d$ individually. The 2D KS test is a generalization of KS test on 2D plane. Although the cumulative probability distribution is not well-defined in more than one dimension, Peacock's insight is that a good surrogate is the integrated probability in each of the four quadrants for a given point $(x,y)$, i.e., $\mathbb{P}(X\leq x,Y\leq y)$, $\mathbb{P}(X\leq x,Y\geq y)$, $\mathbb{P}(X\geq x,Y\leq y)$ and $\mathbb{P}(X\geq x,Y\geq y)$. Similarly, a potential drift is confirmed if the 2D KS test rejects the null hypothesis for at least one of the $d$ bivariate distributions. HHT-AG is summarized in Algorithm \ref{HHT-AG}, where the window size $N$ is set as the number of labeled samples to train the initial classifier $\hat{f}$.

\setlength{\textfloatsep}{10pt plus 1.0pt minus 20.0pt}
\begin{algorithm}[htb]
\small
\caption{\small{HHT with Attribute-wise Goodness of fit (HHT-AG)}}
\label{HHT-AG}
\begin{algorithmic}[1]
\Require
Unlabeled stream $\{\mathbf{X}_t\}_{t=0}^\infty$ where $\mathbf{X}_t\in \mathbb{R}^d$; Significance level $\Theta_1$; Significance level $\Theta_2$ ($=\Theta_1$ by default); Window size $N$.
\Ensure
Detected drift time index $\{T_{cd}\}$; Potential drift time index $\{T_{pot}\}$.
\For {$i = 1$ to $d$}
\State $c_0\leftarrow 0$;
\State $\text{W}_{1,i}\leftarrow$first $N$ points in $x^i$ from time $c_0$;
\State $\text{W}_{2,i}\leftarrow$next $N$ points in $x^i$ in stream;
\EndFor
\While {not end of stream}
\For {$i = 1$ to $d$}
\State Slide $\text{W}_{2,i}$ by $1$ point;
\State Perform KS test with $\Theta_1$ on $\text{W}_{1,i}$ and $\text{W}_{2,i}$;
\If {(KS test rejects the null hypothesis)}
\State $\{T_{pot}\}\leftarrow$current time;
\State $\text{W}_{1}\leftarrow$first $N$ tuples in $(x^i,y)$ from time $c_0$;
\State $\text{W}_{2}\leftarrow$next $N$ tuples in $(x^i,y)$ in stream;
\State Perform 2D KS test with $\Theta_2$ on $\text{W}_{1}$ and $\text{W}_{2}$;
\If {(2D KS test rejects the null hypothesis)}
\State $c_0\leftarrow$current time;
\State $\{T_{cd}\}\leftarrow$current time;
\State Clear all windows and \textbf{GOTO} Step $1$;
\EndIf
\EndIf
\EndFor
\EndWhile
\end{algorithmic}
\normalsize
\end{algorithm}

\section{Experiments} \label{experiments}
Two sets of experiments are performed to evaluate the performance of HHT-CU and HHT-AG. First, quantitative metrics and plots are presented to demonstrate HHT-CU and HHT-AG's effectiveness and superiority over state-of-the-art approaches on benchmark synthetic data. Then, we validate, via three real-world applications, the effectiveness of the proposed HHT-CU and HHT-AG on streaming data classification and the accuracy of its detected concept drift points. This paper selects soft margin SVM as the baseline classifier because of its accuracy and robustness.

\subsection{Experimental Setup} \label{section5.1}
We compare the results with three baseline methods, three topline supervised methods, and two state-of-the-art unsupervised methods for concept drift detection. The first two baselines, DDM~\cite{gama2004learning} and EDDM~\cite{Baena2006Early}, are the most popular supervised drift detector. The third one, we refer to as Attribute-wise KS test (A-KS)~\cite{vzliobaite2010change,reis2016fast}, is a benchmark unsupervised drift detector that has been proved effective in real applications. Note that, A-KS is equivalent to the Layer-I test of HHT-AG. The toplines selected for comparison are LFR~\cite{wang2015concept}, HLFR~\cite{yu2017concept} and HDDM~\cite{frias2015online}. HLFR is the first method on concept drift detection with HHT framework, whereas HDDM introduces Hoeffding's inequality on concept drift detection. All of these methods are operated in supervised manner and significantly outperform DDM. However, LFR and HLFR can only support binary classification. In addition, we also compare with MD3~\cite{sethi2017reliable} and CDBD~\cite{lindstrom2011drift}, the state-of-the-art concept drift detectors that attempt to model $\mathbb{P}_t(y|\ve{X})$ without access to $y$. We use the parameters recommended in the papers for each competing method. The detailed values on significance levels or thresholds (if there exist) are shown in Table \ref{Tab:parameter}.

\begin{table}[!hbpt]
\small
\begin{center}
\begin{tabular}{cccccc}\hline
Algorithms & Significance levels (or thresholds) \\ \hline
\textbf{HHT-CU} &  $\Theta_1=0.01$, $\Theta_2=0.01$ \\
\textbf{HHT-AG} &  $\Theta_1=0.001$, $\Theta_2=0.001$ \\
A-KS &  $\Theta=0.001$ \\
MD3 &  $\Theta=3$ \\
HLFR &  $\delta_\star=0.01$, $\epsilon_\star=0.00001$, $\eta=0.01$ \\
LFR &  $\delta_\star=0.01$, $\epsilon_\star=0.00001$ \\
DDM &  $\alpha=3$, $\beta=2$ \\
EDDM & $\alpha=0.95$, $\beta=0.9$ \\
HDDM & $\alpha_W=0.005$, $\alpha_D=0.001$ \\
\hline
\end{tabular}
\caption{\small{Parameter settings for all competing algorithms.}}\label{Tab:parameter}
\end{center}
\end{table}

\subsection{Results on Benchmark Synthetic Data} \label{section5.2}

\begin{figure}
\centering
\begin{tabular}{cccc}
\subfigure[] {\includegraphics[width=0.48\linewidth]{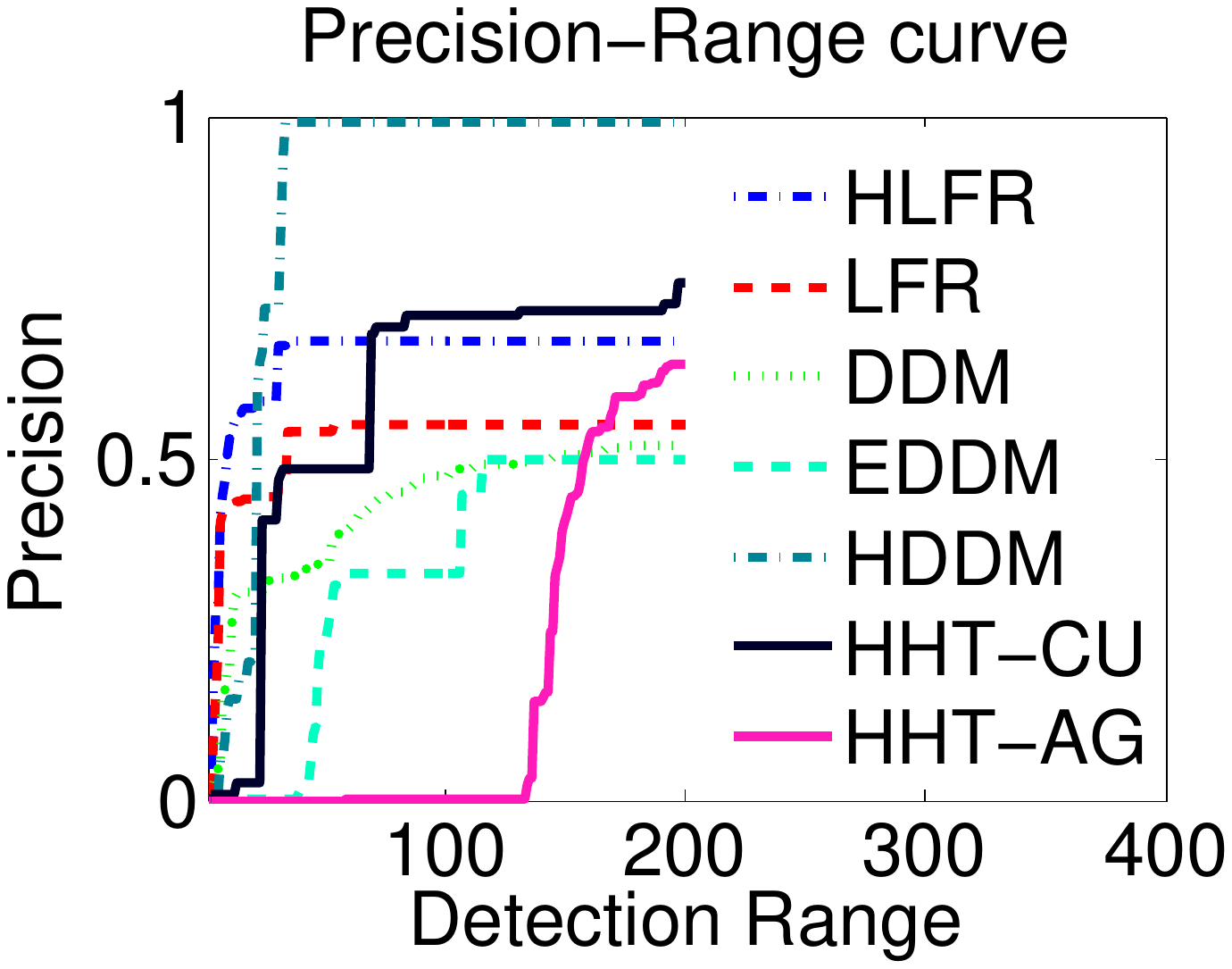}}
\subfigure[] {\includegraphics[width=0.48\linewidth]{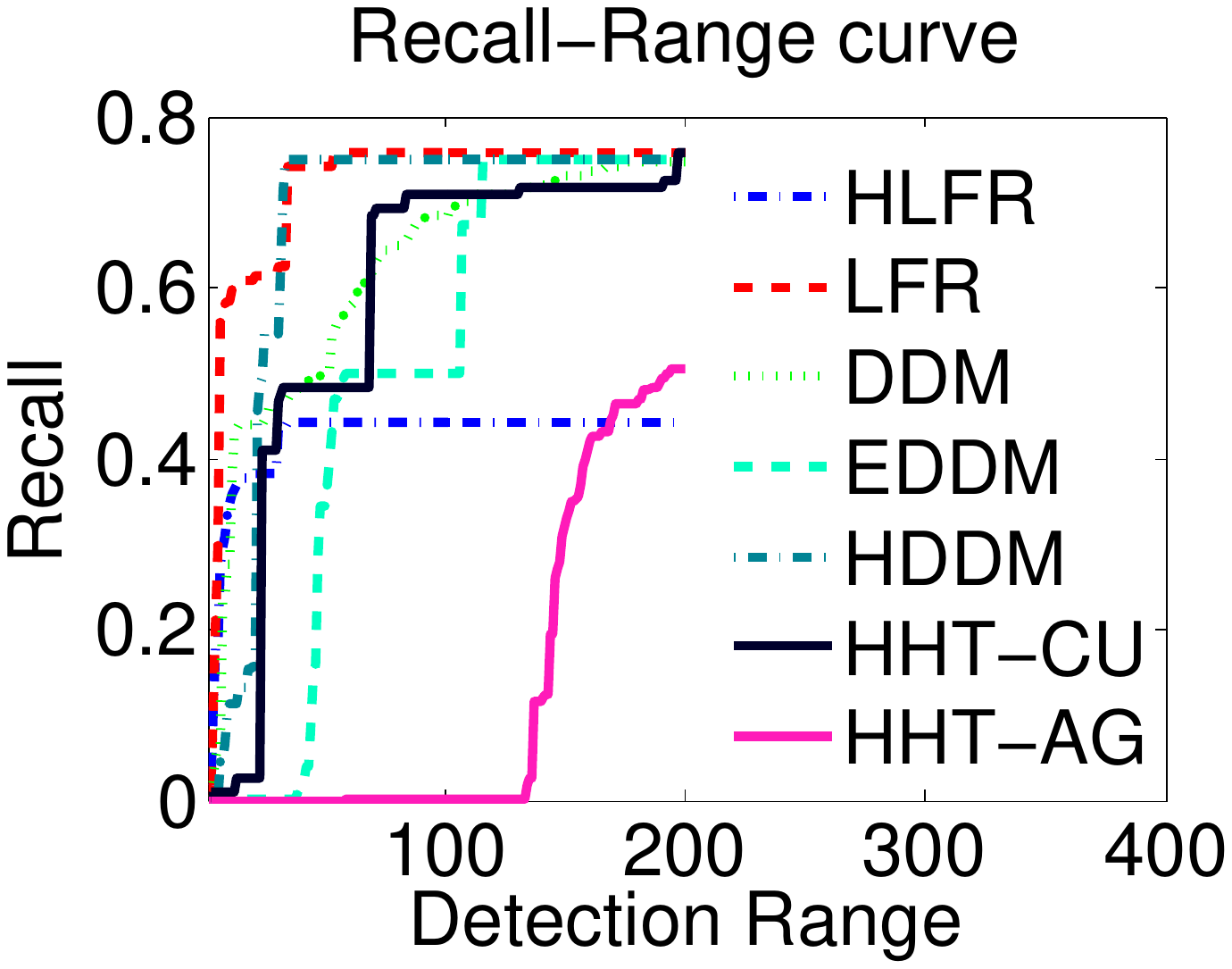}}
\end{tabular}
\caption{\small{Concept drift detection evaluation using (a) the $\mathbf{Precision}$-$\mathbf{Range}$ (PR) curve and (b) the $\mathbf{Recall}$-$\mathbf{Range}$ (RR) curve on UG-2C-2D dataset over $100$ Monte-Carlo trails. The X-axis represents the predefined detection delay range, whereas the Y-axis denotes the corresponding Precision or Recall values. For a specific delay range, a higher $\mathbf{Precision}$ or $\mathbf{Recall}$ value suggests better performance. This figure shows our methods and their supervised counterparts.}}
\label{fig:UG_2C_2D}
\end{figure}

We first compare the performance of the HHT-CU and HHT-AG against aforementioned concept drift approaches on benchmark synthetic data. Eight datasets are selected from \cite{souza2015data,dyer2014compose}, namely 2CDT, 2CHT, UG-2C-2D, MG-2C-2D, 4CR, 4CRE-V1, 4CE1CF, 5CVT. Among them, 2CDT, 2CHT, UG-2C-2D and MG-2C-2D are binary-class datasets, while 4CR, 4CRE-V1, 4CE1CF and 5CVT have multiple classes. To facilitate detection evaluation, we cluster each dataset into $5$ segments to introduce $4$ abrupt drift points, thus controlling ground truth drift points and allowing precise quantitative analysis. Quantitative comparison is performed by evaluating detection quality. To this end, the True Positive ($\emph{TP}$) detection is defined as a detection within a fixed delay range after the precise concept change time. The False Negative ($\emph{FN}$) is defined as missing a detection within the delay range, and the False Positive ($\emph{FP}$) is defined as a detection outside the delay range range or an extra detection in the range. The detection quality is measured jointly with Precision, Recall and delay detection using $\mathbf{Precision}$-$\mathbf{Range}$ curve and $\mathbf{Recall}$-$\mathbf{Range}$ curve respectively (see Fig.~\ref{fig:UG_2C_2D} for an example), where $\mathbf{Precision}=\emph{TP}/(\emph{TP}+\emph{FP})$, and $\mathbf{Recall}=\emph{TP}/(\emph{TP}+\emph{FN})$.

For a straightforward comparison, Table~\ref{Tab:synthetic_labels} reports the number of required labeled samples (in percentage) for each algorithm, whereas Table~\ref{Tab:NAUC} summarizes the Normalized Area Under the Curve (NAUC) values for two kinds of curves. As can be seen, HLFR and LFR can provide the most accurate detection as expected. However, they are only applicable for binary-class datasets and require true labels for the entire data stream. Our proposed HHT-CU and HHT-AG, although slightly inferior to HLFR or LFR, can strike the best tradeoff between detection accuracy and the portion of required labels, especially considering the overwhelming advantage over MD3 and CDBD that are the most relevant counterparts. Although the detection module of MD3 and CDBD are operated in fully unsupervised manner, they either fail to provide reliable detection results or generate too much false positives which may, by contrast, require even more true labels (for classifier update). Meanwhile, it is very encouraging to find that HHT-CU can achieve comparable or even better results than DDM (i.e., the most popular supervised drift detector) with significantly fewer labels. This suggests that our $\emph{classification uncertainty}$ is as sensitive as the total classification accuracy in DDM to monitor the nonstationary environment. And, we can see HHT-AG can significantly improve the Precision value compared to A-KS. This suggests the effectiveness of Layer-II test on reverifying the validity of suspected drifts and denying false alarms. In addition, in the extreme cases when $\mathbb{P}(\ve{X})$ remains unchanged but $\mathbb{P}(y|\ve{X})$ does change, our methods (and the state-of-the-art unsupervised methods) are not able to detect the concept drift which is the change of the joint distribution $\mathbb{P}(\ve{X},y)$. This limitation is demonstrated in our experiments on the synthetic 4CR dataset where $\mathbb{P}(\ve{X})$ remains the same.

\begin{table}[htb]
\small
\setlength{\textfloatsep}{0pt plus 1.0pt minus 10.0pt}
\begin{center}
\begin{tabular}{cccccccccc}\hline
 & HHT-CU & HHT-AG & A-KS & MD3 & CDBD \\ \hline
2CDT & \textcolor{blue}{28.97} & $96.58$ & $78.29$ & \textcolor{red}{13.69} & $96.32$ \\
2CHT & \textcolor{blue}{28.12} & $98.01$ & $77.44$ & \textcolor{red}{11.71} & $93.19$ \\
UG-2C-2D & \textcolor{blue}{28.43} & $37.37$ & \textcolor{red}{18.68} & 31.21 & $88.02$ \\
MG-2C-2D & $25.51$ & $45.46$ & \textcolor{blue}{21.36} & \textcolor{red}{11.83} & $83.58$ \\
4CR &  $0$ & $0$ & $0$ \\
4CRE-V1 &  \textcolor{blue}{29.15} & $40.44$ & \textcolor{red}{20.22} \\
4CR1CF &  \textcolor{blue}{12.69} & $33.64$ & \textcolor{red}{8.33} \\
5CVT &  \textcolor{red}{34.65} & \textcolor{blue}{35.98} & $44.45$ \\ \hline
\end{tabular}
\caption{\small{Averaged number of required labeled samples in the testing set ($\%$) for all competing algorithms. The performances of supervised detectors (HLFR, LFR, DDM, EDDM, and HDDM) are omitted because they require all the true labels (i.e., 100\%). Also, MD3 and CDBD cannot be applied to multi-class datasets including 4CR, 4CRE-V1, 4CR1CF, and 5CVT. The least and second least labeled samples used for each dataset are marked with \textcolor{red}{red} and \textcolor{blue}{blue} respectively. Our methods and A-KS do not detect any drifts on 4CR, and thus they use ``0'' labeled samples for 4CR. MD3 in many cases uses the least labels, but its detection accuracy is the worst as in Table~\ref{Tab:NAUC}.}}\label{Tab:synthetic_labels}
\end{center}
\end{table}
\setlength{\textfloatsep}{10pt plus 1.0pt minus 50.0pt}

\newcolumntype{Y}{>{\centering\arraybackslash}X}
\begin{table*}[!hbpt]
\small
\begin{tabularx}{\textwidth}{@{}lYYYYYYYYYYY@{}}
\toprule
&\multicolumn{2}{c}{\bfseries Our methods}
&\multicolumn{3}{c}{\bfseries Unsupervised methods}
&\multicolumn{3}{c}{\bfseries Supervised methods} \\
\cmidrule(lr){2-3} \cmidrule(l){4-6} \cmidrule(l){7-11}
& HHT-CU & HHT-AG & A-KS & MD3 & CDBD & HLFR & LFR & DDM & EDDM & HDDM \\
\midrule
2CDT &  $\textcolor{red}{0.92}/\textcolor{red}{0.92}$ & $0.15/0.48$ & $0.13/0.62$ & $0.02/0.01$ & $0.08/\textcolor{green}{0.85}$ & $\textcolor{green}{0.82}/0.79$ & $0.81/0.80$ & $0.79/0.79$ & $0.77/0.77$ & $\textcolor{blue}{0.91}/\textcolor{blue}{0.91}$ \\
2CHT &  $\textcolor{green}{0.86}/0.86$ & $0.15/0.48$ & $0.15/0.49$ & $0.02/0.01$ & $0.09/\textcolor{blue}{0.91}$ & $\textcolor{red}{0.93}/\textcolor{red}{0.93}$ & $\textcolor{red}{0.93}/\textcolor{red}{0.93}$ & $0.60/0.60$ & $\textcolor{blue}{0.89}/\textcolor{green}{0.89}$ & $\textcolor{blue}{0.89}/\textcolor{green}{0.89}$\\
UG-2C-2D &  $\textcolor{green}{0.58}/0.58$  & $0.16/0.13$ & $0.08/0.05$ & $0.01/0.07$ & $0.04/\textcolor{red}{0.87}$ & $\textcolor{blue}{0.64}/0.42$ & $0.52/\textcolor{blue}{0.72}$ & $0.43/0.62$ & $0.33/0.49$ & $\textcolor{red}{0.88}/\textcolor{green}{0.67}$\\
MG-2C-2D &  $\textcolor{green}{0.52}/0.52$ & $0.26/0.49$ & $0.21/0.43$ & $0.05/0.16$ & $0.02/\textcolor{blue}{0.80}$ & $\textcolor{red}{0.74}/\textcolor{green}{0.74}$ & $0.46/\textcolor{red}{0.91}$ & $0.37/0.60$ & $0.34/0.73$ & $\textcolor{blue}{0.68}/0.73$\\
4CR &  $-/0$ & $-/0$ & $-/0$ & & & & & $\textcolor{blue}{0.94}/\textcolor{blue}{0.94}$ & $\textcolor{green}{0.86}/\textcolor{green}{0.86}$ & $\textcolor{red}{0.98}/\textcolor{red}{0.98}$ \\
4CRE-V1 &  $\textcolor{green}{0.78}/\textcolor{green}{0.78}$ & $0.21/0.21$ & $0.19/0.21$ & & & & & $0.20/0.22$ & $\textcolor{blue}{0.84}/\textcolor{blue}{0.84}$ & $\textcolor{red}{0.98}/\textcolor{red}{0.98}$ \\
4CR1CF &  $\textcolor{green}{0.49}/0.49$ & $\textcolor{blue}{0.66}/\textcolor{blue}{0.86}$ & $0.43/0.45$ & & & & & $0.10/0.50$ & $0.35/\textcolor{green}{0.63}$ & $\textcolor{red}{0.89}/\textcolor{red}{0.89}$ \\
5CVT &  $\textcolor{red}{0.53}/\textcolor{green}{0.73}$ & $0.16/\textcolor{blue}{0.75}$ & $0.16/\textcolor{red}{0.84}$ & & & & & $\textcolor{blue}{0.43}/\textcolor{green}{0.73}$ & $0.28/0.65$ & $\textcolor{green}{0.35}/0.41$ \\
\bottomrule
\end{tabularx}
\caption{\small{Averaged Normalized Area Under the Curve (NAUC) values for $\mathbf{Precision}$-$\mathbf{Range}$ curve (left side of the forward slash) and $\mathbf{Recall}$-$\mathbf{Range}$ curve (right side of the forward slash) of all competing algorithms. A higher value indicates better performance. The best three results are marked with \textcolor{red}{red}, \textcolor{blue}{blue} and \textcolor{green}{green} respectively. ``-'' denotes no concept drift is detected. MD3, CDBD, HLFR and LFR cannot be applied to multi-class datasets including 4CR, 4CRE-V1, 4CR1CF, and 5CVT. In general, we can see the proposed methods overwhelmingly outperform the unsupervised methods, and achieve similar performances of the supervised methods. In addition, attention should be paid on 4CR, in which only DDM, EDDM, and HDDM can provide satisfactory detection results. This suggests that purely monitoring \emph{classification uncertainty} or modeling the marginal distribution $\mathbb{P}(\ve{X})$ become invalid when there is no change on $\mathbb{P}(\ve{X})$. In this case, sufficient ground truth labels are the prerequisite for reliable detection.}}\label{Tab:NAUC}
\end{table*}

\subsection{Results on Real-world Data} \label{section5.3}

In this section, we evaluate algorithm performance on real-world streaming data classification in a non-stationary environment. Three widely used real-world datasets are selected, namely \textbf{USENET1}~\cite{katakis2008ensemble}, \textbf{Keystroke}~\cite{souza2015data} and \textbf{Posture}~\cite{kaluvza2010agent}. The descriptions on these three datasets are available in \cite{yu2017concept,reis2016fast}. For each dataset, we also select the same number of labeled instances to train the initial classifier as suggested in \cite{yu2017concept,reis2016fast}.

The concept drift detection results and streaming classification results are summarized in Table~\ref{Tab:real_metric}. We measure the cumulative classification accuracy and the portion of required labels to evaluate prediction quality. Since the classes are balanced, the classification accuracy is also a good indicator. In these experiments, our proposed HHT-CU and HHT-AG always feature significantly less amount of false positives, while maintaining good true positive rate for concept drift detection. This suggests the effectiveness of the proposed hierarchical architecture on concept drift reverification. The HHT-CU can achieve overall the best performance in terms of accurate drift detection, streaming classification, as well as the rational utilization of labeled data.

\section{Conclusion}

This paper presents a novel Hierarchical Hypothesis Testing (HHT) framework with a \textbf{Request-and-Reverify} strategy to detect concept drifts. Two methods, namely HHT with Classification Uncertainty (HHT-CU) and HHT with Attribute-wise ``Goodness-of-fit'' (HHT-AG), are proposed respectively under this framework. Our methods significantly outperform the state-of-the-art unsupervised counterparts, and are even comparable or superior to the popular supervised methods with significantly fewer labels. The results indicate our progress on using far fewer labels to perform accurate concept drift detection. The HHT framework is highly effective in deciding label requests and validating detection candidates.

\setlength{\textfloatsep}{10pt plus 1.0pt minus 30.0pt}
\begin{table}[htbp]
  \small
  \centering
  \subtable[\small{USENET1}]{
   \begin{tabular}{cccccccccc}\hline
 & Precision & Recall & Delay & Accuracy & Labels \\ \hline
\textbf{HHT-CU} & $\mathbf{1.00}$ & $\mathbf{1.00}$ & $13.25$ & $\mathbf{85}$ & $\mathbf{30.77}$ \\
\textbf{HHT-AG} & - & 0 & - & $57$ & $0$ \\
A-KS & - & 0 & - & $57$ & $0$ \\
MD3 & $0.14$ & $0.25$ & $16$ & $76$ & $71.85$ \\
CDBD &  $0.10$ & $0.75$ & $\mathbf{3.33}$ & $82$ & $91.15$ \\
HLFR &  $0.75$ & $0.75$ & $11.67$ & $84$ & $100$ \\
LFR &  $0.75$ & $0.75$ & $11.67$ & $84$ & $100$ \\
DDM &  $0.75$ & $0.75$ & $18.33$ & $83$ & $100$ \\
EDDM &  $\mathbf{1.00}$ & $\mathbf{1.00}$ & $57.25$ & $81$ & $100$ \\
HDDM &  $\mathbf{1.00}$ & $\mathbf{1.00}$ & $17.75$ & $83$ & $100$ \\
NA &  - & 0 & - & $57$ & $0$ \\\hline
\end{tabular}
  }
  \subtable[\small{Keystroke}]{
    \begin{tabular}{cccccccccc}\hline
 & Precision & Recall & Delay & Accuracy & Labels \\ \hline
\textbf{HHT-CU} & $\mathbf{1.00}$ & $\mathbf{0.14}$ & $1.5$ & $\mathbf{88}$ & $\mathbf{14.29}$ \\
\textbf{HHT-AG} & $0.5$ & $\mathbf{0.14}$ & $\mathbf{1}$ & $81$ & $57.11$ \\
A-KS & $0.25$ & $\mathbf{0.14}$ & $\mathbf{1}$ & $79$ & $52.43$ \\
DDM &  - & 0 & - & $67$ & $100$ \\
EDDM &  0.33 & 0 & - & $68$ & $100$ \\
HDDM &  $\mathbf{1.00}$ & $\mathbf{0.14}$ & $\mathbf{1}$ & $86$ & $100$ \\
NA &  - & 0 & - & $56$ & $0$ \\ \hline
\end{tabular}
  }
  \subtable[\small{Posture}]{
    \begin{tabular}{cccccccccc}\hline
 & Precision & Recall & Delay & Accuracy & Labels \\ \hline
\textbf{HHT-CU} & $\mathbf{1.00}$ & $\mathbf{1.00}$ & $2421.8$ & $\mathbf{56}$ & $14.60$ \\
\textbf{HHT-AG} & $\mathbf{1.00}$ & $\mathbf{1.00}$ & $\mathbf{406}$ & $55$ & $17.97$ \\
A-KS & $\mathbf{1.00}$ & $\mathbf{1.00}$ & $\mathbf{406}$ & $55$ & $\mathbf{10.54}$ \\
DDM &  $0.75$ & $0.75$ & $3318.67$ & $54$ & $100$ \\
EDDM &  $0.75$ & $0.75$ & $1253.4$ & $54$ & $100$ \\
HDDM &  $\mathbf{1.00}$ & $\mathbf{1.00}$ & $689.25$ & $\mathbf{56}$ & $100$ \\
NA &  - & 0 & - & $46$ & $0$ \\ \hline
\end{tabular}
  }
  \caption{\small{Quantitative metrics on real-world applications. The $\mathbf{Precision}$, $\mathbf{Recall}$ and $\mathbf{Delay}$ denote the concept drift detection precision value, recall value and detection delay, whereas the $\mathbf{Accuracy}$ and $\mathbf{Labels}$ denote the cumulative classification accuracy and required portion of true labels in the testing set (\%). ``-'' denotes no concept drift is detected or the detected drift points all are false alarms. ``NA'': using initial classifier without any update.}}\label{Tab:real_metric}
\end{table}
\setlength{\textfloatsep}{10pt plus 1.0pt minus 10.0pt}

\nocite{gonccalves2014comparative,sobolewski2013concept,brzezinski2014prequential,wang2013concept}
\bibliographystyle{named}
\bibliography{refbib_reduce}

\end{document}